\documentclass[conference]{IEEEtran}
\IEEEoverridecommandlockouts
\usepackage{booktabs}
\usepackage{cite}
\usepackage{amsmath,amssymb,amsfonts}
\usepackage{algorithmic}
\usepackage{graphicx}
\usepackage{textcomp}
\usepackage{xcolor}
\def\BibTeX{{\rm B\kern-.05em{\sc i\kern-.025em b}\kern-.08em
    T\kern-.1667em\lower.7ex\hbox{E}\kern-.125emX}}
\begin{document}

\title{Multimodal AI for Body Fat Estimation: Computer Vision and Anthropometry with DEXA Benchmarks}

\author{\IEEEauthorblockN{Rayan ALDajani}
\IEEEauthorblockA{\textit{Dept. of Electrical Engineering and Computer Science} \\
\textit{York University}\\
Toronto, Canada \\
rayandajani21@gmail.com}
}

\maketitle

\begin{abstract}
Tracking body fat percentage is essential for effective weight loss and health management, yet gold-standard methods such as DEXA scans [1], [2] are too expensive and rarely accessible for most people. This study aims to evaluate the feasibility of artificial intelligence (AI) models as low-cost alternatives using frontal body images and basic anthropometric data. The dataset consists of 535 samples: 253 cases with recorded anthropometric measurements (weight, height, neck, ankle, and wrist) and 282 images obtained via web scraping from Reddit posts self-reported body fat percentage values, some of which were stated to be derived from official DEXA scans. As no publicly available datasets exist for computer vision based body fat estimation, this dataset was compiled specifically for this study. Two approaches were developed: (1) ResNet-based image models, (2) regression models using measurements only. A multimodal fusion approach was proposed but could not be implemented due to the lack of paired datasets, and is identified as future work. The image-based model achieved a Root Mean Square Error (RMSE) of 4.44\% and a Coefficient of Determination ($R^{2}$) of 0.807. These results show that AI-assisted models and tools can give low-cost and accessible body fat estimates. This supports a future of consumer based weight loss and fitness apps.
\end{abstract}


\section{Introduction}
\subsection{The Importance of Accessible Body Fat Estimation}

Body fat percentage (BF\%) is an increasingly popular marker of health, fitness, and metabolic risk. Although accurate BF\% tracking is important for measuring weight loss, athletic performance, and disease prevention, existing validated methods are largely unavailable to the general public or too unreliable for weekly tracking. To enable widespread use in health and fitness, estimation must be both accurate and affordable.
\subsection{Limitations of Current Methods}
Dual-Energy X-ray Absorptiometry (DEXA), the gold-standard assessment of body fat [1], [2], provides an accurate body composition assessment and detailed health profiles not obtainable with any other technique. DEXA scans are expensive, time-consuming, and impractical for routine fitness tracking. Low-cost assessments, like the U.S. Navy method [3], which relies on neck, waist, and height measurements, are available for the individual's use at home but are not consistent or accurate enough for appropriate monitoring of fitness and health.

\subsection{Role of Artificial Intelligence}\label{AA}
Recent advances in artificial intelligence (AI) and computer vision offer the possibility of combining affordability with reliability. Deep learning has been applied in obesity detection, pose estimation, and medical imaging [4]. However, no public datasets exist for image-based body fat estimation, leaving a gap in research and practical applications.

\subsection{Contribution of This Study}
This study addresses this gap by providing a new dataset of images and anthropometric data and testing two approaches: (1) ResNet-based regression on the image data only, (2) regression on the measurements only. Our results show that the image-based model yields the best estimates, with respect to DEXA-referenced cases, suggesting the possibility for AI-driven, low-cost consumer applications for monitoring body fat.

\section{Background}
\subsection{Anthropometric Models}
Low fidelity calculations, such as the Body Mass Index (BMI) or the U.S. Navy approach, are commonplace in the literature because they are inexpensive and easy to use; however, they are not very accurate because they do not account for various nuances of body composition.
\subsection{Gold-standard Imaging Techniques}
DEXA provides accurate fat, lean mass, and density measurements, but is expensive and not widely used [1], [2], [5].
\subsection{Artificial Intelligence in Health Applications}
Recent ML and vision advances enable health prediction tasks like disease detection and body composition estimation. These advances offer a cost-effective and scalable approach to DEXA-type measurements.

\section{Results}
\subsection{Dataset Description}
Two datasets were used. The image dataset consisted of 282 frontal body images with self-reported percentages of body fat (some referencing DEXA scans) and the anthropometric dataset consisted of 253 male records (Fisher and Johnson) where weight, height, and body circumferences were documented (body fat obtained from underwater weighing and Siri’s equation). Paired samples were not available, thus each modality was considered independently. For both datasets, an 80/20 train/test split was applied with randomized shuffling and no subject overlap. A validation subset was not used due to dataset size. Early stopping was employed to prevent overfitting.

\begin{table}[h]
\centering
\caption{Cohort characteristics (mean $\pm$ SD).}
\label{tab:cohort_summary}
\begin{tabular}{lcc}
\toprule
Metric & Anthropometric (n=253) & Image (n=282) \\
\midrule
Body Fat (\%)     & 19.1 $\pm$ 8.3  & 20.03 $\pm$ 10.3\\
Height (cm)       & 178.15 $\pm$ 3.66& N/A \\
Weight (kg)       & 80.7 $\pm$ 29.3& N/A \\
Male/Female (\%)  & 100/0 & 100/0  \\
\bottomrule
\end{tabular}
\end{table}

\subsection{Anthropometric Model Results}
The regression model performed moderately well on the test set (MAE = 3.52\%; RMSE = 4.47\%; R² = 0.57). It was trained on five anthropometric measures (weight, chest, abdomen, hip, thigh). Abdominal circumference was the strongest predictor of body fat, consistent with previous findings linking it to central adiposity. Compared to BMI, the regression model showed better predictive performance; however, its error range was too large for reliable standalone use.

\subsection{Image-Based Model Results}
With training on 282 frontal images that reported body fat (43\% DEXA reference), the Convolutional Neural Network (CNN) achieved MAE = 3.34\%, RMSE = 4.44, and R² = 0.81 with test data, outperforming the regression model. The CNN predictions articulated well on mid-range body fat, with larger variability in very low and very high ranges. The training curves indicated stable convergence with only mild overfitting due to the small dataset. Overall, the CNN had increased predictive capability of body fat than BMI and the Navy heuristics, highlighting the need for larger and more diverse data for generalization. Future versions will benchmark lightweight baselines and deeper CNNs to quantify model-choice effects.

\section{Discussion and Conclusion}
The regression model using five anthropometric measures produced reasonable accuracy (R² = 0.57, RMSE = 4.47\%), with abdominal circumference as the most predictive factor. Since the models were trained on different datasets, the comparison is qualitative. Dataset size and difficulty may affect the performance gap. The CNN trained on 282 images achieved better results (R² = 0.81, RMSE = 4.44\%) by capturing visual cues absent from measurements. Both models outperformed heuristics (e.g., BMI, Navy equation) but remained less accurate than DEXA. Future work will test advanced architectures, expand population diversity, and collect paired multimodal samples (image + anthropometry + DEXA). The dataset also serves as a benchmark for privacy-preserving, at-home body-fat estimation. Overall, results support the feasibility of AI-based body fat prediction without the cost of DEXA.
\begin{figure}[ht]
    \centering
    \includegraphics[width=0.31\textwidth]{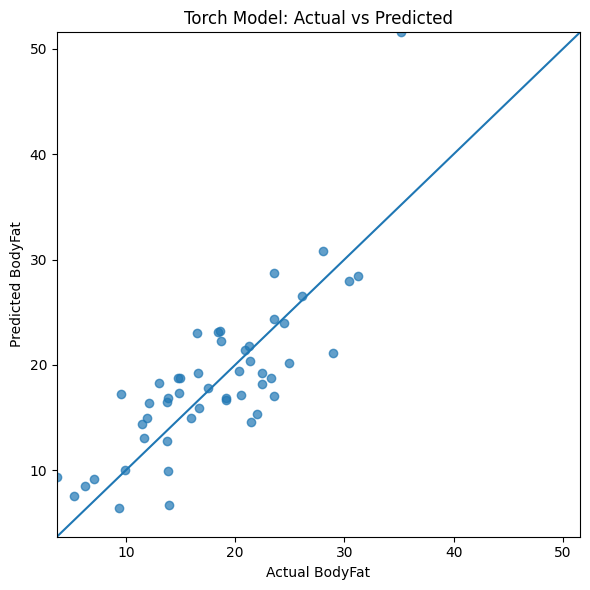}
    \caption{Scatter plot of predicted vs.\ true body fat percentage for the anthropometric regression model (using weight, chest, abdomen, hip, and thigh). The $R^2$ value of 0.57 indicates moderate predictive ability.}
    \label{fig:anthro_scatter}
\end{figure}
\begin{figure}[ht]
    \centering
    \includegraphics[width=0.35\textwidth]{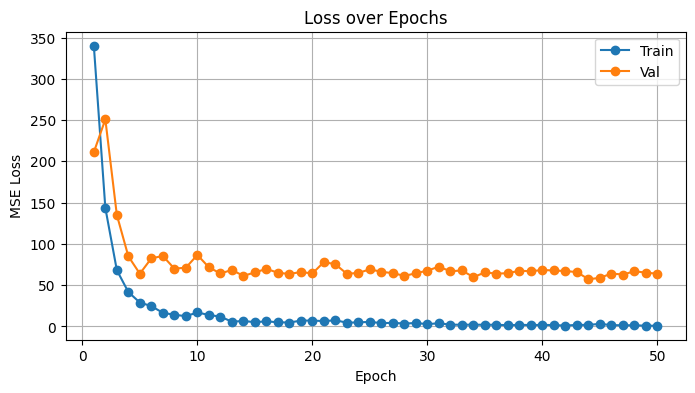}
    \caption{Training and validation loss curves for the image-based CNN model over 50 epochs. The model converges stably, with training loss decreasing steadily and validation loss plateauing at a higher level.}
    \label{fig:oss_over_epoch}
\end{figure}

\end{document}